%% file: acl2020.tex
\documentclass[11pt,a4paper]{article}
\usepackage[hyperref]{acl2020}
\usepackage{times}
\usepackage{latexsym}

\usepackage{microtype}
\usepackage{graphicx}
\usepackage{amsmath}
\usepackage{amssymb}
\usepackage{booktabs} 
\usepackage{tablefootnote}
\usepackage{threeparttable}
\usepackage{enumitem}
\usepackage[table,xcdraw]{}
\usepackage{array}
\usepackage{float}
\usepackage{helvet}

\newcolumntype{C}[1]{%
>{\raggedleft\hspace{0pt}}p{#1}}%

\aclfinalcopy 
\def\aclpaperid{2421} 

\title{Visual Grounding Methods for VQA are Working for the Wrong Reasons!}

\author{Robik Shrestha$^1$ \quad Kushal Kafle$^{1,2}$ \quad Christopher Kanan$^{1,3,4}$ \\
Rochester Institute of Technology$^1$ \quad Adobe Research$^2$ \quad Paige$^3$ \quad Cornell Tech$^4$ \\
\texttt{\{rss9369, kk6055, kanan\}@rit.edu}
}

\date{}

\begin{document}
\maketitle
\begin{abstract}
Existing Visual Question Answering (VQA) methods tend to exploit dataset biases and spurious statistical correlations, instead of producing right answers for the right reasons. To address this issue, recent bias mitigation methods for VQA propose to incorporate visual cues (e.g., human attention maps) to better ground the VQA models, showcasing impressive gains. However, we show that the performance improvements are not a result of improved visual grounding, but a regularization effect which prevents over-fitting to linguistic priors. For instance, we find that it is not actually necessary to provide proper, human-based cues; random, insensible cues also result in similar improvements. Based on this observation, we propose a simpler regularization scheme that does not require any external annotations and yet achieves near state-of-the-art performance on VQA-CPv2\footnotemark.

\end{abstract}

\section{Introduction}

\label{sec:introduction}

Visual Question Answering (VQA)~\cite{antol2015vqa}, the task of answering questions about visual content, was proposed to facilitate the development of models with human-like visual and linguistic understanding. However, existing VQA models often exploit superficial statistical biases to produce responses, instead of producing the right answers for the right reasons~\cite{kafle2019challenges}. 

The VQA-CP dataset~\cite{agrawal2018don} showcases this phenomenon by incorporating different question type/answer distributions in the train and test sets. Since the linguistic priors in the train and test sets differ, models that  exploit these priors fail on the test set. To tackle this issue, recent works have endeavored to enforce proper visual grounding, where the goal is to make models produce answers by looking at relevant visual regions~\cite{gan2017vqs,selvaraju2019taking,wu2019self}, instead of  exploiting linguistic priors. These approaches rely on additional annotations/cues such as human-based attention maps~\cite{das2017human}, textual explanations~\cite{huk2018multimodal} and object label predictions~\cite{ren2015faster} to identify relevant regions, and train the model to base its predictions on those regions, showing large improvements (8-10\% accuracy) on the VQA-CPv2 dataset.
\footnotetext{\url{https://github.com/erobic/negative_analysis_of_grounding}}

\begin{figure}
 \centering
  \footnotesize
    \includegraphics[width=0.98\linewidth]{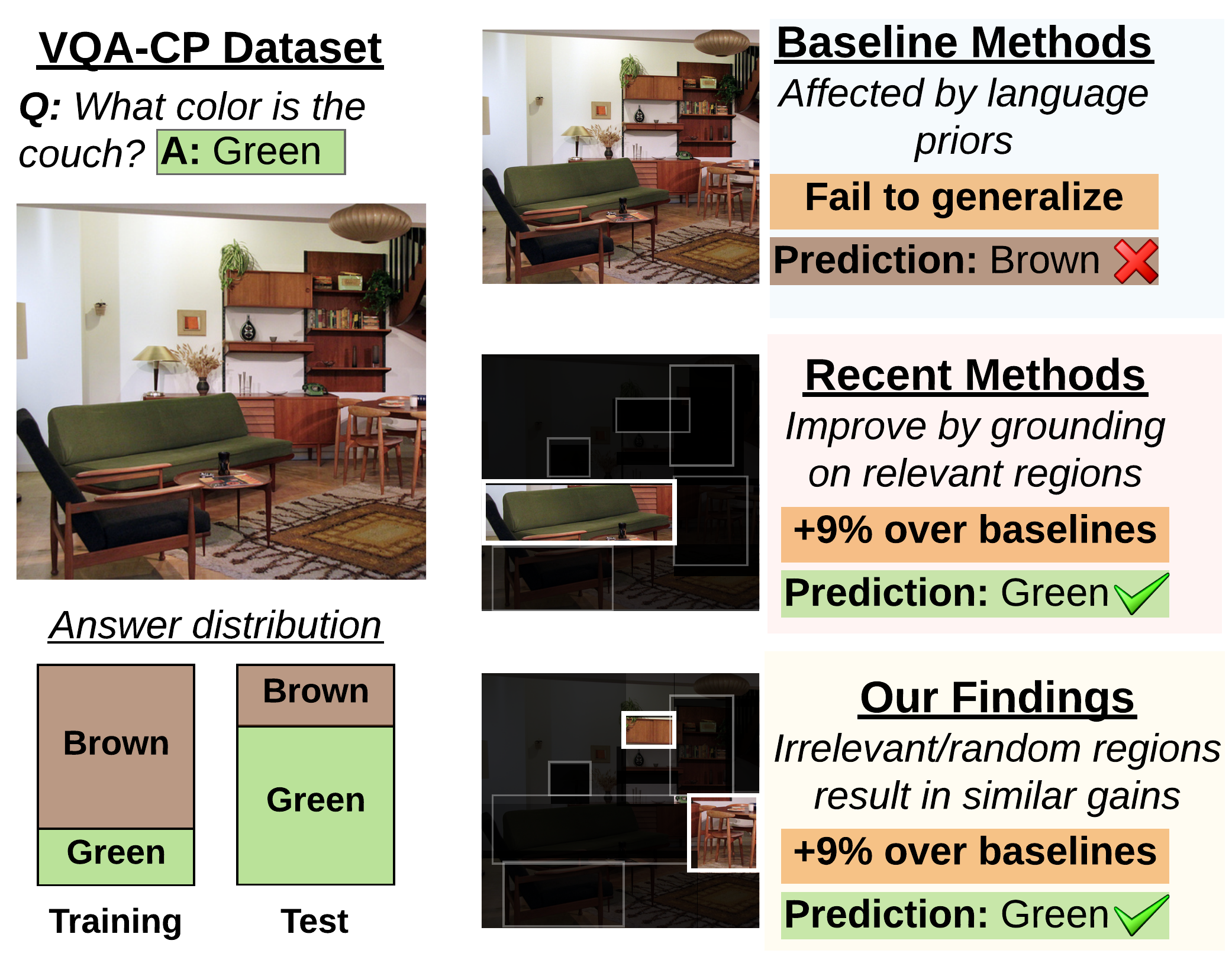}
      \caption{We find that existing visual sensitivity enhancement methods improve performance on VQA-CPv2 through regularization as opposed to proper visual grounding.}
\end{figure}

Here, we study these methods. We find that their improved accuracy does not actually emerge from proper visual grounding, but from regularization effects, where the model \textit{forgets} the linguistic priors in the train set, thereby performing better on the test set. To support these claims, we first show that it is possible to achieve such gains even when the model is trained to look at: a) irrelevant visual regions, and b) random visual regions. Second, we show that differences in the predictions from the variants trained with relevant, irrelevant and random visual regions are not statistically significant. Third, we show that these methods degrade performance when the priors remain intact and instead work on VQA-CPv2 by hurting its train accuracy.

Based on these observations, we hypothesize that controlled degradation on the train set allows models to forget the training priors to improve test accuracy. To test this hypothesis, we introduce a simple regularization scheme that zeros out the ground truth answers, thereby always penalizing the model, whether the predictions are correct or incorrect. We find that this approach also achieves near state-of-the-art performance ($48.9\%$ on VQA-CPv2), providing further support for our claims.

While we agree that visual grounding is a useful direction to pursue, our experiments show that the community requires better ways to test if systems are actually visually grounded. We make some recommendations in the discussion section.

\section{Related Work}

\subsection{Biases in VQA}
As expected of any real world dataset, VQA datasets also contain dataset biases~\cite{goyal2017making}. The VQA-CP dataset~\cite{agrawal2018don} was introduced to study the robustness of VQA methods against linguistic biases. Since it contains different answer distributions in the train and test sets, VQA-CP makes it nearly impossible for the models that rely upon linguistic correlations to perform well on the test set~\cite{agrawal2018don, shrestha2019ramen}.

\subsection{Bias Mitigation for VQA}
VQA algorithms without explicit bias mitigation mechanisms fail on VQA-CP, so recent works have focused on the following solutions:

\subsubsection{Reducing Reliance on Questions} 
Some recent approaches employ a question-only branch as a control model to discover the questions most affected by linguistic correlations. The question-only model is either used to perform adversarial regularization~\cite{grand2019adversarial,ramakrishnan2018overcoming} or to re-scale the loss based on the difficulty of the question~\cite{cadene2019rubi}. However, when these ideas are applied to the UpDn model~\cite{Anderson2017up-down}, which attempts to learn correct visual grounding, these approaches achieve 4-7\% lower accuracy compared to the state-of-the-art methods.

\subsubsection{Enhancing Visual Sensitivities} 
Both Human Importance Aware Network Tuning (HINT)~\cite{selvaraju2019taking} and Self Critical Reasoning (SCR)~\cite{wu2019self}, train the network to be more sensitive towards salient image  regions by improving the alignment between visual cues and gradient-based sensitivity scores. HINT proposes a ranking loss between human-based importance scores~\cite{das2016human} and the gradient-based sensitivities. In contrast, SCR does not require exact saliency ranks. Instead, it penalizes the model if correct answers are more sensitive towards non-important regions as compared to important regions, and if incorrect answers are more sensitive to important regions than correct answers. 

\section{Existing VQA Methods}
Given a question $\mathcal{Q}$ and an image $\mathcal{I}$, \textit{e.g.,} represented by bottom-up region proposals: $v$ ~\cite{Anderson2017up-down}, a VQA model is tasked with predicting the answer $a$:
\begin{align}
    P(a|\mathcal{Q}, \mathcal{I}) = f_{VQA}(v, \mathcal{Q}).
\end{align}

\subsection{Baseline VQA Methods} Without additional regularization, existing VQA models such as the baseline model used in this work: UpDn~\cite{Anderson2017up-down}, tend to rely on the linguistic priors: $P(a|\mathcal{Q})$ to answer questions. Such models fail on VQA-CP, because the priors in the test set differ from the train set. 

\subsection{Visual Sensitivity Enhancement Methods}
To reduce the reliance on linguistic priors, visual sensitivity enhancement methods attempt to train the model to be more sensitive to relevant visual regions when answering questions. Following~\cite{wu2019self}, we define the sensitivity of an answer $a$ with respect to a visual region $v_i$ as:
\begin{align}
    \mathcal{S}(a, v_i) := (\nabla_{v_i} P(a|\mathcal{I}, \mathcal{Q}))^T \mathbf{1}.
\end{align}

Existing methods propose the following training objectives to improve grounding using $\mathcal{S}$:
\begin{itemize}[noitemsep,leftmargin=*]
    \item \textbf{HINT} uses a ranking loss, which penalizes the model if the pair-wise rankings of the sensitivities of visual regions towards ground truth answers $a_{gt}$ are different from the ranks computed from the human-based attention maps.
    
    \item \textbf{SCR} divides the region proposals into influential and non-influential regions and penalizes the model if: 1) $\mathcal{S}(a_{gt})$ of a non-influential region is higher than an influential region, and 2) the region most influential for the correct answer has even higher sensitivity for incorrect answers.
\end{itemize}
Both methods improve baseline accuracy by 8-10\%. Is this actually due to better visual grounding?

\section{Why Did the Performance Improve?}
\label{sec:improvement-reasons}
We probe the reasons behind the performance improvements of HINT and SCR. We first analyze if the results improve even when the visual cues are irrelevant (Sec.~\ref{sec:irrelevant_regions}) or random (Sec.~\ref{sec:random_regions}) and examine if their differences are statistically significant (Sec.~\ref{sec:stats}). Then, we analyze the regularization effects by evaluating the performance on VQA-CPv2's train split (Sec.~\ref{sec:drop_in_train}) and the behavior on a dataset without changing priors (Sec.~\ref{sec:drop_in_vqav2}). We present a new metric to assess visual grounding in Sec.~\ref{sec:cpig} and describe our regularization method in Sec.~\ref{sec:our-approach}.

\subsection{Experimental Setup}
\label{sec:experimental_setup}
We compare the baseline UpDn model with HINT and SCR-variants trained on VQAv2 or VQA-CPv2 to study the causes behind the improvements. We report mean accuracies across $5$ runs, where a pre-trained UpDn model is fine-tuned on subsets with human attention maps and textual explanations for HINT and SCR respectively. Further training details are provided in the Appendix.

\subsection{Training on Irrelevant Visual Cues}
\label{sec:irrelevant_regions}

In our first experiment we studied how irrelevant visual cues performed compared to relevant ones. We fine-tune the model with irrelevant cues defined as: $\mathcal{S}_{irrelevant} := (1 - \mathcal{S}_h)$, where, $\mathcal{S}_h$ represents the human-based importance scores. As shown in the `Grounding using irrelevant cues' section of Table~\ref{tab:main-results}, both HINT and SCR are within 0.3\% of the results obtained from looking at relevant regions, which indicates  the gains for HINT and SCR are not necessarily from looking at relevant regions.

\input{tables/main-results.tex}

\subsection{Training on Random Visual Cues}
\label{sec:random_regions}

In our next experiment we studied how random visual cues performed with HINT and SCR.  We assign random importance scores to the visual regions: $\mathcal{S}_{rand} \sim
 \textit{uniform}(0,1)$. We test two variants of randomness: \textbf{Fixed random regions}\textit{}, where $\mathcal{S}_{rand}$ are fixed once chosen, and \textbf{Variable random regions}, where $\mathcal{S}_{rand}$ are regenerated every epoch. As shown in Table~\ref{tab:main-results}, both of these variants obtain similar results as the model trained with human-based importance scores. The performance improves even when the importance scores are changed every epoch, indicating that it is not even necessary to look at the \textit{same} visual regions.
 
\subsection{Significance of Statistical Differences}
\label{sec:stats}
To test if the changes in results were statistically significant, we performed Welch's t-tests~\cite{welch1938significance} on the predictions of the variants trained on relevant, irrelevant and random cues. We pick Welch's t-test over the Student's t-test, because the latter assumes equal variances for predictions from different variants. 

To perform the tests, we first randomly sample $5000$ subsets of non-overlapping test instances. We then average the accuracy of each subset across $5$ runs, obtaining $5000$ values. Next, we run the t-tests for HINT and SCR separately on the subset accuracies. As shown in Table~\ref{tab:hint-overlaps}, the $p$-values across the variants of HINT and SCR are greater than or equal to $0.3$. Using a confidence level of $95\%$ ($\alpha = 0.05$), we fail to reject the null hypothesis that the mean difference between the paired values is $0$, showing that the variants are not statistically significantly different from each other. We also compare the predictions of HINT/SCR against baseline, and find that $p$-values are all zeros, showing that the differences have statistical significance.

\input{tables/statistical-tests.tex}
\textbf{Percentage of Overlaps:}  To further check if the variants trained on irrelevant or random regions gain performance in a manner similar to the models trained on relevant regions, we compute the overlap between their predictions on VQA-CPv2's test set. The percentage of overlap is defined as:
\begin{align*}
    \%~Overlap = \frac{n_{same}}{n_{total}} \times 100\%,
\end{align*}
where, $n_{same}$ denotes the number of instances where either both variants were correct or both were incorrect and $n_{total}$ denotes the total number of test instances. As shown in Table~\ref{tab:hint-overlaps}, we compare $\%Overlap$ between different variants of HINT/SCR with baseline and against each other. We find $89.7-91.9\%$ and $89.5-92.0\%$ overlaps for different variants of HINT and SCR respectively. These high overlaps suggest that the variants are not working in fundamentally different manners.

\subsection{Drops in Training Accuracy}
\label{sec:drop_in_train}
We compare the training accuracies to analyze the regularization effects. As shown in Table~\ref{tab:main-results}, the baseline method has the highest training results, while the other methods cause $6.0-14.0\%$ and $3.3-10.5\%$ drops in the training accuracy on VQA-CPv2 and VQAv2, respectively. We hypothesize that degrading performance on the train set helps forget linguistic biases, which in turn helps accuracy on VQA-CPv2's test set but hurts accuracy on VQAv2's val set.

\subsection{Drops in VQAv2 Accuracy}
\begin{figure}
 \footnotesize
    \includegraphics[width=0.9\linewidth]{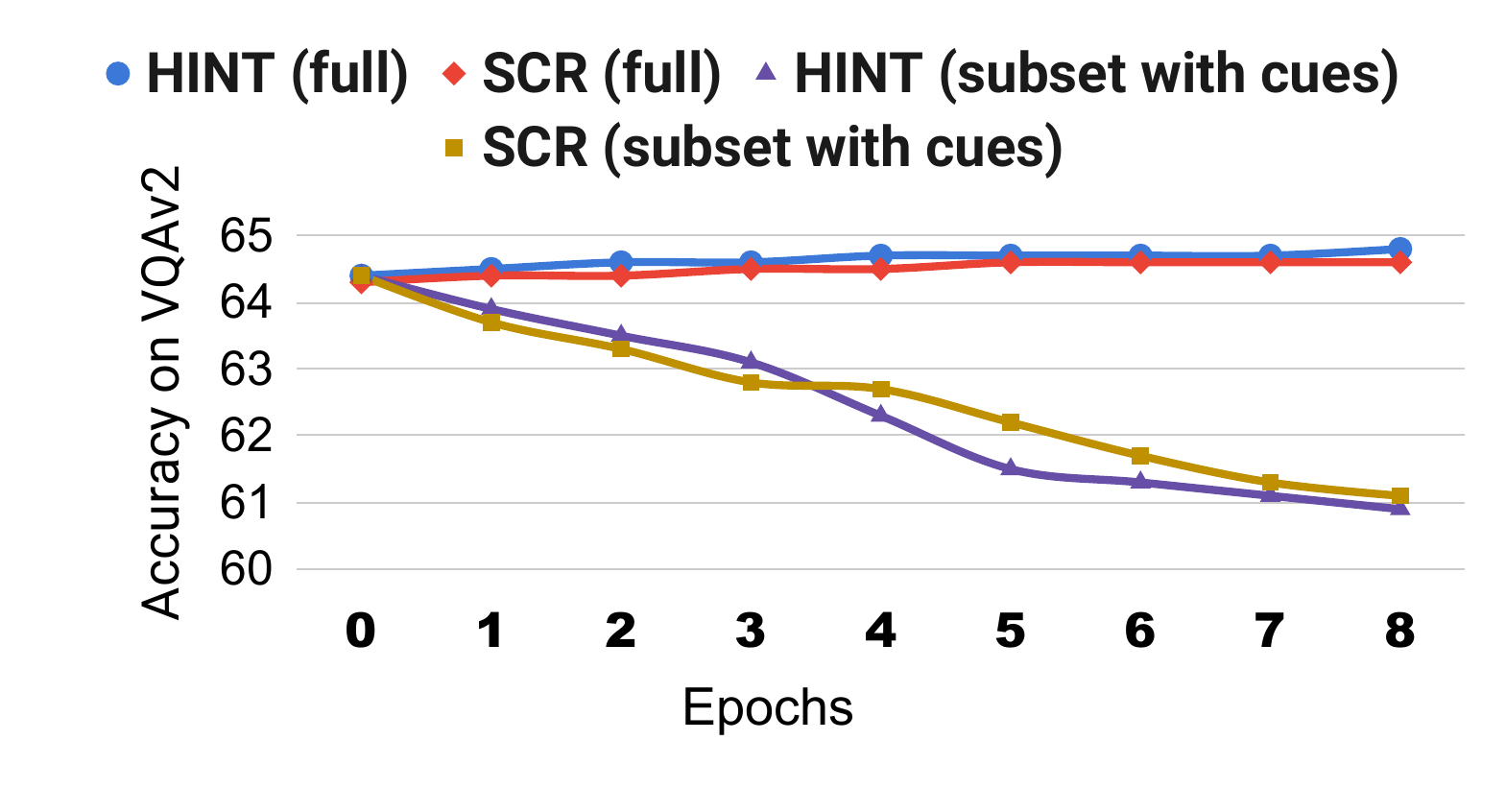}
      \caption{Accuracies for HINT and SCR on VQAv2's val set, when fine-tuned either on the full train set or on the subset containing visual cues.}
      \label{fig:drop_in_vqav2}
\end{figure}
\label{sec:drop_in_vqav2}
As observed by~\citet{selvaraju2019taking} and as shown in Fig.~\ref{fig:drop_in_vqav2}, we observe small improvements on VQAv2 when the models are fine-tuned on the entire train set. However, if we were to compare against the improvements in VQA-CPv2 in a fair manner, i.e., only use the instances with visual cues while fine-tuning, then, the performance on VQAv2 drops continuously during the course of the training. This indicates that HINT and SCR help forget linguistic priors, which is beneficial for VQA-CPv2 but not for VQAv2.
 
\subsection{Assessment of Proper Grounding}
\label{sec:cpig}
In order to quantitatively assess visual grounding, we propose a new metric called: Correctly Predicted but Improperly Grounded (CPIG):
\begin{align*}
    \% CPIG = \frac{N_{\text{correct ans, improper grounding}}}{N_{\text{correct ans}}} \times 100\%,
\end{align*}
which is the number instances for which the most sensitive visual region used to correctly predict the answer is not within top-3 most relevant ground truth regions, normalized by the total number of correct predictions. HINT and SCR trained on relevant regions obtained lower CPIG values that other variants (70.24\% and 80.22\% respectively), indicating they are better than other variants at finding relevant regions. However, these numbers are still high, and show that only 29.76\% and 19.78\% of the correct predictions for HINT and SCR were properly grounded. Further analysis is presented in the Appendix.

\section{Embarrassingly Simple Regularizer}
\label{sec:our-approach}
The usage of visual cues and sensitivities in existing methods is superfluous because the results indicate that performance improves through degradation of training accuracy. We hypothesize that simple regularization that does not rely on cues or sensitivities can also achieve large performance gains for VQA-CP. To test this hypothesis, we devise a simple loss function which continuously degrades the training accuracy by training the network to always predict a score of zero for all possible answers i.e. produce a zero vector ($\mathbf{0}$). The overall loss function can be written as:
\begin{align*}
    L := BCE(P(\mathcal{A}), \mathcal{A}_{gt}) + \lambda BCE(P(\mathcal{A}), \mathbf{0}),
\end{align*} 
where, BCE refers to the binary cross entropy loss and $P(\mathcal{A})$ is a vector consisting of predicted scores for all possible answers. The first term is the binary cross entropy loss between model predictions and ground truth answer vector ($\mathcal{A}_{gt}$), and the second term is our regularizer with a coefficient of $\lambda=1$. Note that this regularizer continually penalizes the model during the course of the training, whether its predictions are correct or incorrect. 

As shown in Table~\ref{tab:main-results}, we present results when this loss is used on: a) Fixed subset covering $1\%$ of the dataset, b) Varying subset covering $1\%$ of the dataset, where a new random subset is sampled every epoch and c) $100\%$ of the dataset. Confirming our hypothesis, all variants of our model achieve near state-of-the-art results, solidifying our claim that the performance gains for recent methods come from regularization effects. 

It is also interesting to note that the drop in training accuracy is lower with this regularization scheme as compared to the state-of-the-art methods. Of course, if any model was actually visually grounded, then we would expect it to improve performances on both train and test sets. We do not observe such behavior in any of the methods, indicating that they are not producing right answers for the right reasons.

\section{Discussion on Proper Grounding}

While our results indicate that current visual grounding based bias mitigation approaches do not suffice, we believe this is still a good research direction. However, future methods must  seek to verify that performance gains are not stemming from spurious sources by using an experimental setup similar to that presented in this paper.  We recommend that both train and test accuracy be reported, because a model truly capable of visual grounding would not cause drastic drops in training accuracy to do well on the test sets. Finally, we advocate for creating a dataset with ground truth grounding available for 100\% of the instances using synthetically generated datasets~\cite{kafle2017data,kafle2017analysis,kafle2018dvqa,acharya2019tallyqa,Hudson2019GQAAN,johnson2017clevr}, enabling the community to evaluate if their methods are able to focus on relevant information. Another alternative is to use tasks that explicitly test grounding, e.g., in visual query detection an agent must output boxes around any regions of a scene that match the natural language query~\cite{acharya2019vqd}.

\section{Conclusion}

Here, we showed that existing visual grounding based bias mitigation methods for VQA are not working as intended. We found that the accuracy improvements stem from a regularization effect rather than proper visual grounding. We proposed a simple regularization scheme which, despite not requiring additional annotations, rivals state-of-the-art accuracy. Future visual grounding methods should be tested with a more comprehensive experimental setup and datasets for proper evaluation.

\textbf{Acknowledgement.} 
This work was supported in part by AFOSR grant [FA9550-18-1-0121], NSF award \#1909696, and a gift from Adobe Research. We thank NVIDIA for the GPU donation. The views and conclusions contained herein are those of the authors and should not be interpreted as representing the official policies or endorsements of any sponsor. We are grateful to Tyler Hayes for agreeing to review the paper at short notice and suggesting valuable edits and corrections for the paper.

\bibliography{acl2020}
\bibliographystyle{acl_natbib}

\input{supplementary}

\end{document}

%% file: tables/main-results.tex
\begin{table}[t]
\caption{Results on VQA-CPv2 and VQAv2 datasets for the baseline UpDn, visual sensitivity enhancement methods (HINT and SCR) and our own regularization method, including the published (pub.) numbers.\label{tab:main-results}}
\footnotesize
\begin{threeparttable}
\begin{tabular}{lrrrr}
\toprule
                   \multicolumn{1}{C{0.5cm}}{} & \multicolumn{2}{C{1.8cm}}{\textbf{VQA-CPv2}} & \multicolumn{2}{C{1.8cm}}{\textbf{VQAv2}} \\ \midrule
                   & \multicolumn{1}{C{0.9cm}}{Train}         & \multicolumn{1}{C{0.9cm}}{Test}         & \multicolumn{1}{C{0.9cm}}{Train}         & \multicolumn{1}{C{0.9cm}}{Val}        \\ \midrule \midrule
\multicolumn{5}{l}{\textit{Baseline - Without visual grounding}}              \\  
UpDn               & 84.0  &             40.1 &               83.4 &       64.4     \\ \midrule 
\multicolumn{5}{l}{\textit{Grounding using human-based cues}}             \\ 
HINT$_{pub.}$          & N/A    &              46.7 &             N/A &     63.4\tnote{1}
          \\
SCR$_{pub.}$                & N/A              & \textbf{49.5}              & N/A              & 62.2           \\
HINT               &   73.9            & 48.2             & 75.7              &    61.3        \\
SCR                &    75.9           &  49.1            &  77.9            & 61.3            \\ 
\midrule 
\multicolumn{5}{l}{\textit{Grounding using irrelevant cues}}     \\ 
HINT  &    71.2          & 48.0              &               73.5 & 60.3           \\
SCR  &      75.7         &  49.2            &              74.1 & 59.1            \\ \midrule 

\multicolumn{5}{l}{\textit{Grounding using fixed random cues}}     \\ 
HINT & 72.0              & 48.1              & 73.0               & 59.5            \\
SCR  &   70.0            & 49.1             & 78.0               & 61.4            \\ \midrule 

\multicolumn{5}{l}{\textit{Grounding using variable random cues}}     \\ 
HINT &   71.9            & 48.1             & 72.9               & 59.4           \\
SCR  &  69.6             & 49.2              & 78.1              & 61.5           \\ \midrule

\multicolumn{5}{l}{\textit{Regularization by zeroing out answers}}     \\ 
Ours$_{1\%~fixed}$ &   \textbf{78.0}            & 48.9             & \textbf{80.1}               & \textbf{62.6}           \\
Ours$_{1\%~var.}$ &   77.6            & 48.5             & 80.0               & \textbf{62.6}           \\
Ours$_{100\%}$ &   75.7            & 48.2             & 79.9               & 62.4           \\
         \bottomrule
\end{tabular}
\begin{tablenotes}
    \small
    \item[1] The published number is a result of fine-tuning HINT on the entire training set, but as described in Sec.~\ref{sec:drop_in_vqav2}, other published numbers and our experiments fine-tune only on the instances with cues.
\end{tablenotes}


\end{threeparttable}
\end{table}

%% file: tables/statistical-tests.tex
\begin{table}
\caption{$p$-values from the Welch's t-tests and the percentage of overlap between the predictions  (Ovp.) of different variants of HINT and SCR.\label{tab:hint-overlaps}}
\footnotesize
\begin{tabular}{lrr}
\toprule
                                         Methods & $p$ & Ovp.(\%) \\
\midrule                                          
\multicolumn{3}{l}{\textit{HINT variants against Baseline}} \\
\midrule                  
\midrule                  
Default vs. Baseline               & 0.0 & 83.6           \\
Irrelevant vs. Baseline            & 0.0 & 82.4           \\
Fixed Random vs. Baseline          & 0.0 & 82.0           \\
Variable Random vs. Baseline       & 0.0 & 81.5           \\
\midrule                                          
\midrule                  
\multicolumn{3}{l}{\textit{Among HINT variants}}                \\
\midrule                  
Default vs Irrelevant       & 0.3 & 89.7           \\
Default vs Fixed random     & 0.7 & 90.9           \\
Default vs Variable random  & 0.6 &  91.9           \\
Irrelevant vs Fixed random        & 0.5 &  95.6           \\
Irrelevant vs Variable random   & 0.7 & 93.9      \\
Fixed random vs Variable random   & 0.9 & 96.9 \\
\midrule
\midrule     
\multicolumn{3}{l}{\textit{SCR variants against Baseline}} \\
\midrule                  
\midrule                  
Default vs. Baseline               & 0.0 & 85.6           \\
Irrelevant vs. Baseline            & 0.0 & 84.2           \\
Fixed Random vs. Baseline          & 0.0 & 80.7           \\
Variable Random vs. Baseline       & 0.0 & 80.6           \\
\midrule                                          
\midrule                  
\multicolumn{3}{l}{\textit{Among SCR variants}}                \\
\midrule                  
Default vs Irrelevant       & 0.6 & 92.0           \\
Default vs Fixed random     & 0.8 & 89.3           \\
Default vs Variable random      & 0.6 &  89.5           \\
Irrelevant vs Fixed random        & 0.4 &  91.7           \\
Irrelevant vs Variable random   & 1.0 & 91.6      \\
Fixed random vs Variable random   & 0.4 & 96.7 \\
\midrule
\end{tabular}

\end{table}

%% file: supplementary.tex
\newcolumntype{C}[1]{%
>{\raggedleft\hspace{0pt}}p{#1}}%

\renewcommand{\thepage}{A\arabic{page}}  
\renewcommand{\thesection}{A}   
\renewcommand{\thetable}{A\arabic{table}}   
\renewcommand{\thefigure}{A\arabic{figure}}

\section{Appendix}
\subsection{Training Details}

We compare four different variants of HINT and SCR to study the causes behind the improvements including the models that are fine-tuned on: 1) relevant regions (state-of-the-art methods) 2) irrelevant regions 3) fixed random regions and 4) variable random regions. For all variants, we fine-tune a pre-trained UpDn, which was trained on either VQA-CPv2 or VQAv2 for 40 epochs with a learning rate of $10^{-3}$. When fine-tuning with HINT, SCR or our method, we also use the main binary cross entropy VQA loss, whose weight is set to $1$. The batch size is set to $384$ for all of the experiments.

\subsubsection*{HINT} Following~\cite{selvaraju2019taking}, we train HINT on the subset with human-based attention maps~\cite{das2017human}, which are available for 9\% of the VQA-CPv2 train and test sets. The same subset is used for VQAv2 too. The learning rate is set to $2 \times 10^{-5}$ and the weight for the HINT loss is set to $2$.

\subsubsection*{SCR} Since~\cite{wu2019self} reported that human-based textual explanations~\cite{huk2018multimodal} gave better results than human-based attention maps for SCR, we train all of the SCR variants on the subset containing textual explanation-based cues. SCR is trained in two phases. For the first phase, which strengthens the influential objects, we use a learning rate of $5 \times 10^{-5}$, loss weight of $3$ and train the model to a maximum of 12 epochs. Then,~following~\cite{wu2019self}, for the second phase, we use the best performing model from the first phase to train the second phase, which criticizes incorrect dominant answers. For the second phase, we use a learning rate of $10^{-4}$ and weight of $1000$, which is applied alongside the loss term used in the first phase. The specified hyperparameters worked better for us than the values provided in the original paper.

\subsubsection*{Our Zero-Out Regularizer} Our regularization method, which is a binary cross entropy loss between the model predictions and a zero vector, does not use additional cues or sensitivities and yet achieves near state-of-the-art performance on VQA-CPv2. We set the learning rate to: $\frac{2 \times 10^{-6}}{r}$, where $r$ is the ratio of the training instances used for fine-tuning. The weight for the loss is set to $2$. We report the performance obtained at the $8^{th}$ epoch.

\subsection{Results}

\input{tables/rank-correlation.tex}

\subsubsection*{Correlation with Ground Truth Visual Cues} 
Following~\cite{selvaraju2019taking}, we report Spearman's rank correlation between network's sensitivity scores and  human-based scores in Table~\ref{tab:rank-correlation}. For HINT and our zero-out regularizer, we use human-based attention maps. For SCR, we use textual explanation-based scores. We find that HINT trained on human attention maps has the highest correlation coefficients for both datasets. However, compared to baseline, HINT variants trained on random visual cues also show improved correlations. For SCR, we obtain surprising results, with the model trained on irrelevant cues obtaining higher correlation than that trained on relevant visual cues. As expected, applying our regularizer does not improve rank correlation. Since HINT trained on relevant cues obtains the highest correlation values, it does indicate improvement in visual grounding. However, as we have seen, the improvements in performance cannot necessarily be attributed to better overlap with ground truth localizations.

\subsubsection*{A Note on Qualitative Examples}

\begin{figure*}
 \centering
    \includegraphics[width=0.9\textwidth]{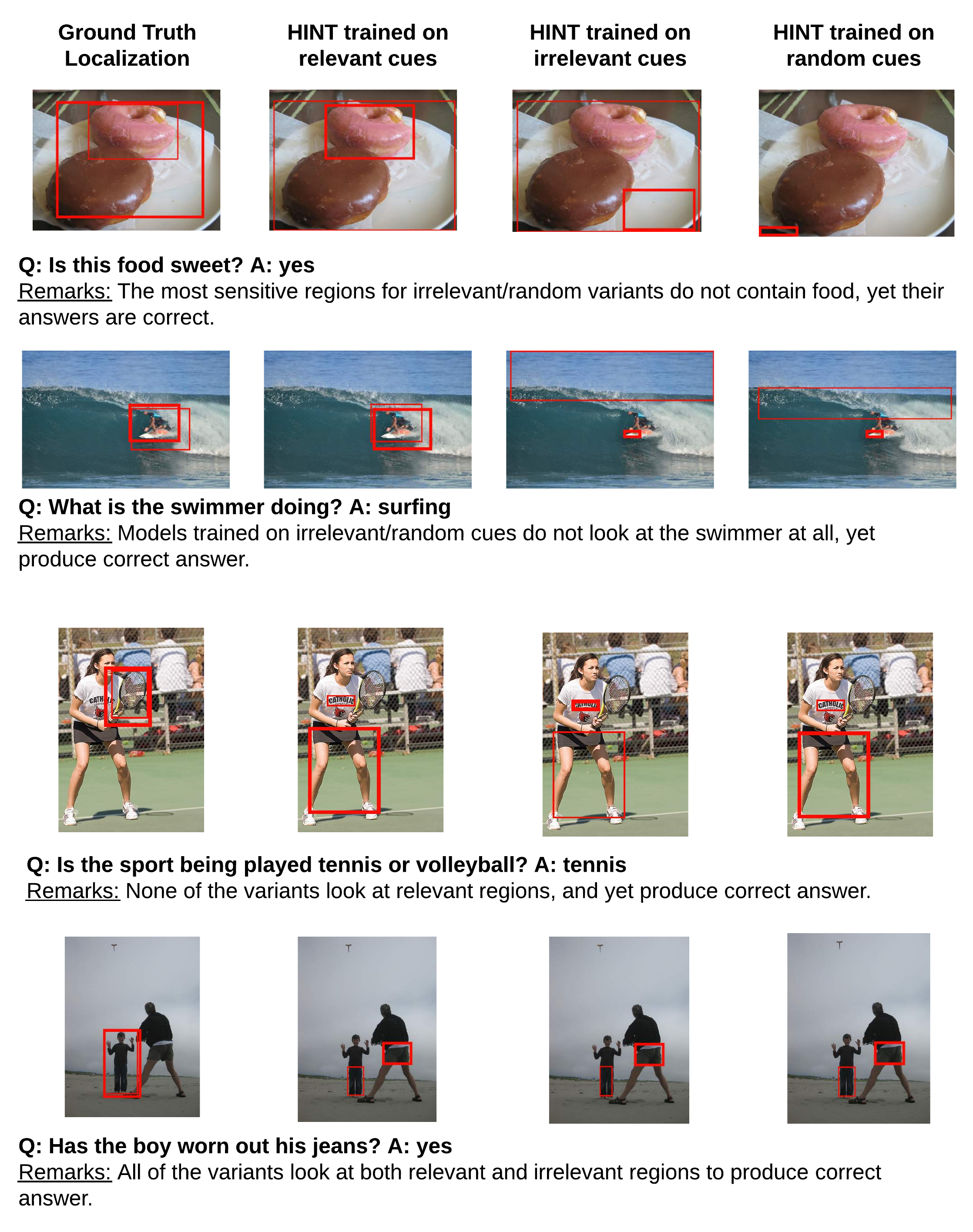}\caption{Visualizations of most sensitive visual regions used by different variants of HINT to make predictions. We pick samples where all variants produce correct response to the question. The first column shows ground truth regions and columns 2-4 show visualizations from HINT trained on relevant, irrelevant and fixed random regions respectively.  \label{fig:grad-cam}}
\end{figure*}

Presentation of qualitative examples in visual grounding models for VQA suffers from confirmation bias i.e., while it is possible to find qualitative samples that look at relevant regions to answer questions properly, it is also possible to find samples that produce correct answers without looking at relevant regions. We present examples for such cases in Fig.~\ref{fig:grad-cam}. We next present a quantitative assessment of visual grounding, which does not suffer from the confirmation bias.

\subsubsection*{Quantitative Assessment of Grounding} 
\begin{figure*}
 \centering
    \includegraphics[width=0.9\textwidth]{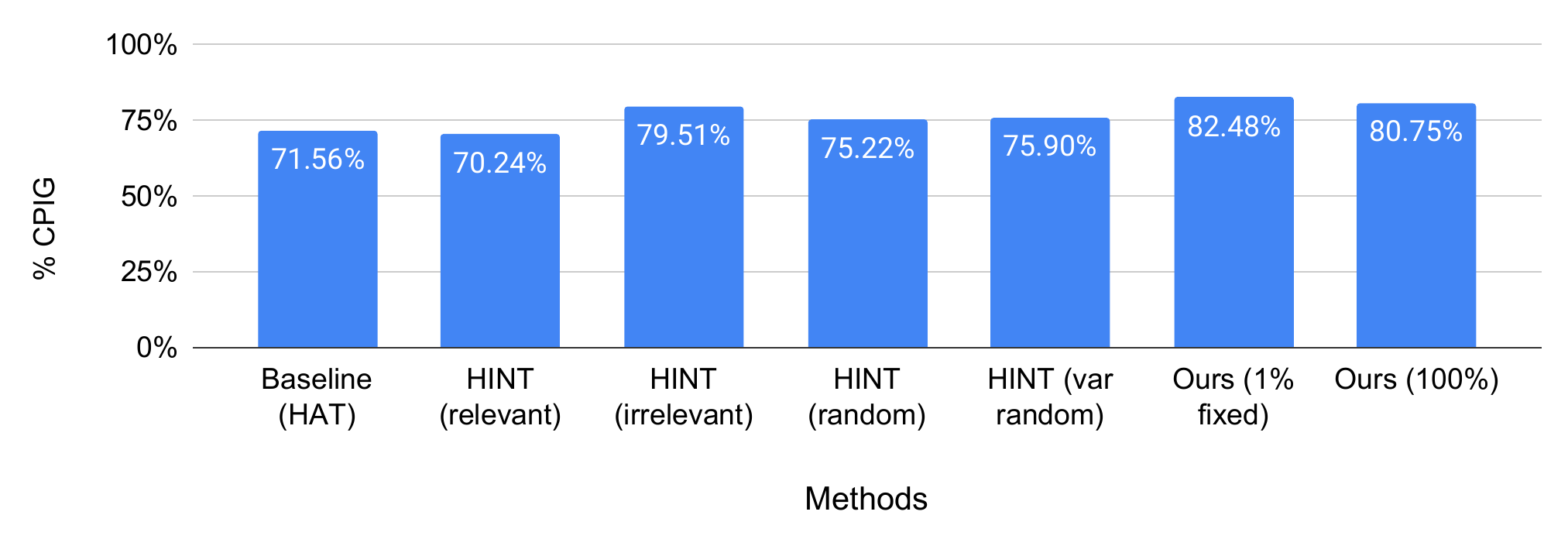}
      \caption{\% CPIG for baseline and different variants of HINT and our method, computed using ground truth relevant regions taken from human attention maps (lower is better).\label{fig:HINT-cpig}}
\end{figure*}

\begin{figure*}
 \centering
    \includegraphics[width=0.9\textwidth]{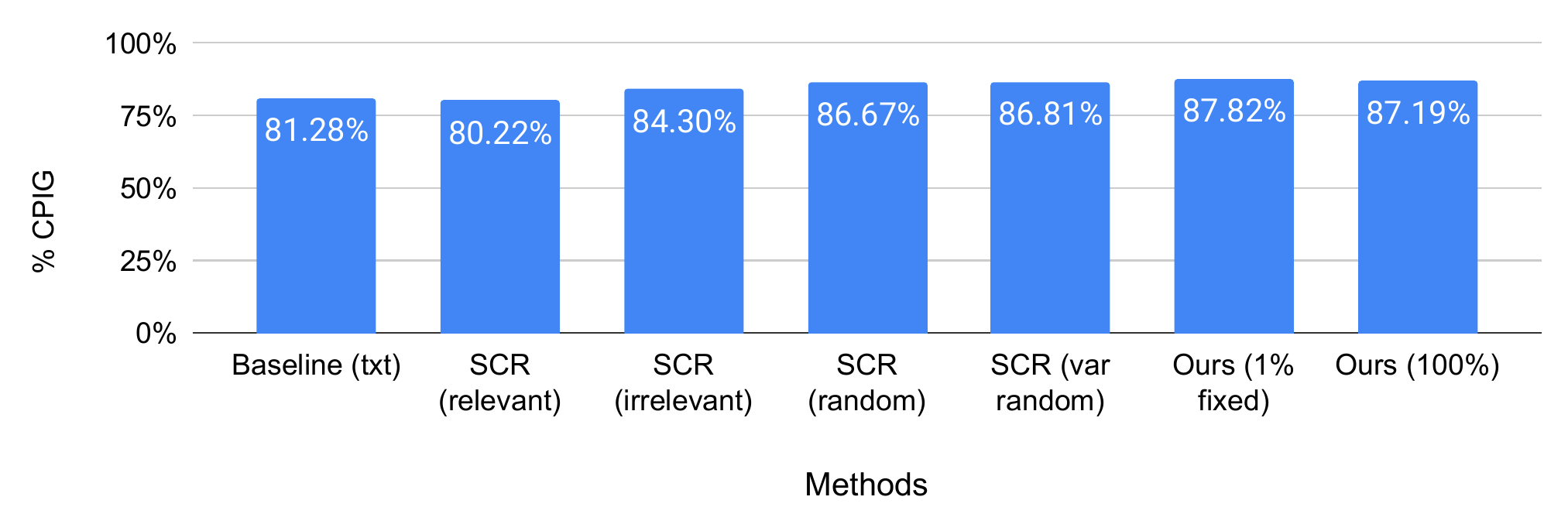}
      \caption{\% CPIG for baseline and different variants of SCR and our method, computed using ground truth relevant regions taken from textual explanations (txt)\label{fig:SCR-cpig}.}
\end{figure*}

In order to truly assess if existing methods are using relevant regions to produce correct answers, we use our proposed metric: Correctly Predicted but Improperly Grounded (CPIG). If the CPIG values are large, then it implies that large portion of correctly predicted samples were not properly grounded. Fig.~\ref{fig:HINT-cpig} shows $\%$ CPIG for different variants of HINT trained on human attention-based cues, whereas Fig.~\ref{fig:SCR-cpig} shows the metric for different variants of SCR trained on textual explanation-based cues. We observe that HINT and SCR trained on relevant regions have the lowest $\%$ CPIG values (70.24\% and 80.22\% respectively), indicating that they are better than other variants in finding relevant regions. However, only a small percentage of correctly predicted samples were properly grounded (29.76\% and 19.78\% for HINT and SCR respectively), even when trained on relevant cues.


\subsubsection*{Breakdown by Answer Types}
Table~\ref{tab:score-per-answer-type} shows VQA accuracy for each answer type on VQACPv2's test set. HINT/SCR and our regularizer show large gains in `Yes/No' questions. We hypothesize that the methods help forget linguistic priors, which improves test accuracy of such questions. In the train set of VQACPv2, the answer `no' is more frequent than the answer `yes', tempting the baseline model to answer `yes/no' questions with `no'. However, in the test set, answer `yes' is more frequent. Regularization effects caused by HINT/SCR and our method cause the models to weaken this prior i.e., reduce the tendency to just predict `no', which would increase accuracy at test because `yes' is more frequent in the test set. Next, all of the methods perform poorly on `Number (Num)' answer type, showing that methods find it difficult to answer questions that are most reliant on correct visual grounding such as: localizing and counting objects. Finally, we do not observe large improvements in `Other' question type, most likely due to the large number of answers present under this answer type.
\input{tables/score-per-answer-type.tex}

\begin{figure*}
 \centering
    \includegraphics[width=0.9\textwidth]{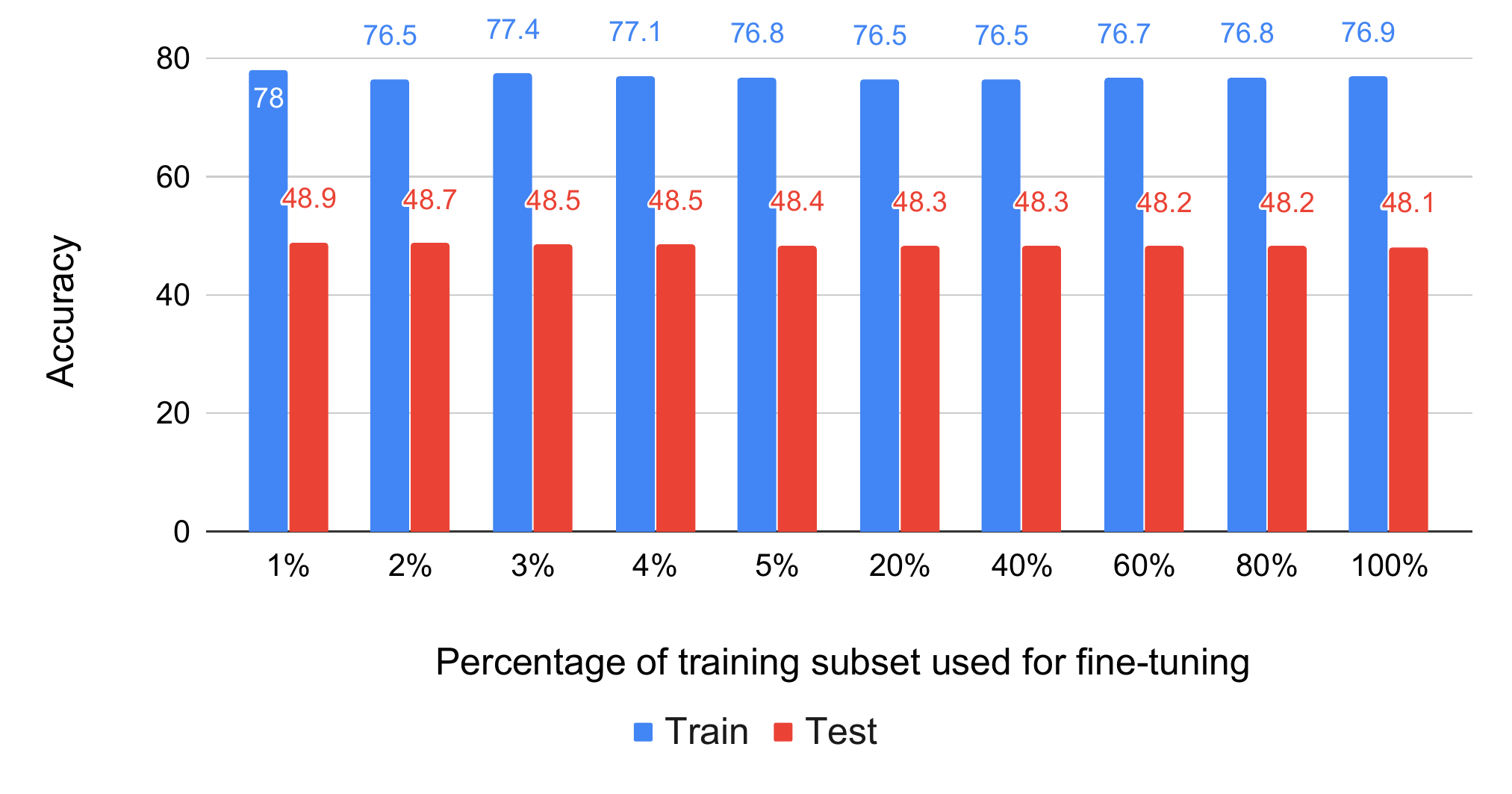}
      \caption{The regularization effect of our loss is invariant with respect to the dataset size.\label{fig:acc-vs-dsize}}
\end{figure*}
\subsubsection*{Accuracy versus Size of Train Set} We test our regularization method on random subsets of varying sizes. Fig.~\ref{fig:acc-vs-dsize} shows the results when we apply our loss to $1 - 100 \%$ of the training instances. Clearly, the ability to regularize the model does not vary much with respect to the size of the train subset, with the best performance occurring when our loss is applied to $1\%$ of the training instances. These results support our claims that it is possible to improve performance without actually performing visual grounding.

%% file: tables/rank-correlation.tex
\begin{table}[t]
\caption{Results on VQA-CPv2 and VQAv2 datasets for the baseline UpDn, visual sensitivity enhancement methods (HINT and SCR) and our own regularization method, including the published (pub.) numbers.\label{tab:rank-correlation}}
\footnotesize
\begin{threeparttable}
\begin{tabular}{lrrrr}
\toprule
                   \multicolumn{1}{C{1.5cm}}{} & \multicolumn{1}{C{2cm}}{\textbf{VQA-CPv2}} & \multicolumn{1}{C{2cm}}{\textbf{VQAv2}} \\ \midrule
                   \midrule
\multicolumn{3}{l}{\textit{Baseline - Without visual grounding}}              \\  
UpDn               & 0.0110 &       0.0155     \\ \midrule
\multicolumn{3}{l}{\textit{Grounding using human-based cues}}             \\ 
HINT          & 0.1020  &     0.1350\\
SCR         & 0.0340  & -0.0670 \\
\midrule 
\multicolumn{3}{l}{\textit{Grounding using irrelevant cues}}     \\ 
HINT  &    -0.0048          & -0.0200 \\
SCR  &  0.0580 & -0.0100 \\ \midrule 

\multicolumn{3}{l}{\textit{Grounding using fixed random cues}}     \\ 
HINT & 0.0510              & 0.0620 \\
SCR  & -0.0250            & -0.0350 \\ \midrule 

\multicolumn{3}{l}{\textit{Grounding using variable random cues}}     \\ 
HINT &   0.0570            & 0.0623 \\
SCR  &  -0.0380             & 0.0246  \\ \midrule

\multicolumn{3}{l}{\textit{Regularization by zeroing out answers}}     \\ 
Ours$_{1\%~fixed}$ &   -0.1050            & -0.1200 \\
Ours$_{100\%}$ &  -0.0750 & -0.0100  \\
         \bottomrule
\end{tabular}

\end{threeparttable}
\end{table}

%% file: tables/score-per-answer-type.tex
\begin{table}[t]
\caption{VQA accuracy per answer-type on VQACPv2 test set.\label{tab:score-per-answer-type}}
\footnotesize
\begin{threeparttable}
\begin{tabular}{lrrrr}
\toprule
                   \multicolumn{1}{C{1.5cm}}{} & \multicolumn{1}{C{0.9cm}}{Overall}         & \multicolumn{1}{C{0.9cm}}{Yes/No}         & \multicolumn{1}{C{0.9cm}}{Num}         & \multicolumn{1}{C{0.9cm}}{Other}        \\ \midrule \midrule
\multicolumn{5}{l}{\textit{Baseline - Without visual grounding}}              \\  
UpDn               & 40.1  &    41.1          &               12.0 &       47.2     \\ \midrule 
\multicolumn{5}{l}{\textit{Grounding using human-based cues}}             \\ 
HINT               &   48.2            & 65.2             & 13.8              &    47.5        \\
SCR                &    49.1           &  70.3            &  11.5            & \textbf{48.0}            \\ 
\midrule 
\multicolumn{5}{l}{\textit{Grounding using irrelevant cues}}     \\ 
HINT  &    48.0          & 67.2               &               13.5 & 47.1           \\
SCR  &      \textbf{49.2}         & 73.4              &              11.5 & 46.4            \\ \midrule 

\multicolumn{5}{l}{\textit{Grounding using fixed random cues}}     \\ 
HINT & 48.1              & 66.9              & 13.8               & 46.9            \\
SCR  &   49.1            & \textbf{74.7}             & 12.2               & 45.1            \\ \midrule 

\multicolumn{5}{l}{\textit{Grounding using variable random cues}}     \\ 
HINT &   48.1            & 67.1             & \textbf{13.9}               & 46.9           \\
SCR  &  49.2             & \textbf{74.7}              & 12.2              & 45.1           \\ \midrule

\multicolumn{5}{l}{\textit{Regularization by zeroing out answers}}     \\ 
Ours$_{1\%~fixed}$ &  48.9            & 69.8             & 11.3               & 47.8           \\
Ours$_{100\%}$ &   48.2            & 66.7             & 11.7               & 47.9           \\
         \bottomrule
\end{tabular}
\end{threeparttable}
\end{table}

%% file: acl2020.bbl
\begin{thebibliography}{24}
\expandafter\ifx\csname natexlab\endcsname\relax\def\natexlab#1{#1}\fi

\bibitem[{Acharya et~al.(2019{\natexlab{a}})Acharya, Jariwala, and Kanan}]{acharya2019vqd}
Manoj Acharya, Karan Jariwala, and Christopher Kanan. 2019{\natexlab{a}}.
\newblock {VQD}: Visual query detection in natural scenes.
\newblock In \emph{Proceedings of the 2019 Conference of the North {A}merican Chapter of the Association for Computational Linguistics: Human Language Technologies, Volume 1 (Long and Short Papers)}, pages 1955--1961, Minneapolis, Minnesota. Association for Computational Linguistics.

\bibitem[{Acharya et~al.(2019{\natexlab{b}})Acharya, Kafle, and Kanan}]{acharya2019tallyqa}
Manoj Acharya, Kushal Kafle, and Christopher Kanan. 2019{\natexlab{b}}.
\newblock Tallyqa: Answering complex counting questions.
\newblock In \emph{Association for the Advancement of Artificial Intelligence (AAAI)}.

\bibitem[{Agrawal et~al.(2018)Agrawal, Batra, Parikh, and Kembhavi}]{agrawal2018don}
Aishwarya Agrawal, Dhruv Batra, Devi Parikh, and Aniruddha Kembhavi. 2018.
\newblock Don’t just assume; look and answer: Overcoming priors for visual question answering.
\newblock In \emph{Proceedings of the IEEE Conference on Computer Vision and Pattern Recognition (CVPR)}, pages 4971--4980.

\bibitem[{Anderson et~al.(2018)Anderson, He, Buehler, Teney, Johnson, Gould, and Zhang}]{Anderson2017up-down}
Peter Anderson, Xiaodong He, Chris Buehler, Damien Teney, Mark Johnson, Stephen Gould, and Lei Zhang. 2018.
\newblock Bottom-up and top-down attention for image captioning and visual question answering.
\newblock In \emph{Proceedings of the IEEE Conference on Computer Vision and Pattern Recognition (CVPR)}.

\bibitem[{Antol et~al.(2015)Antol, Agrawal, Lu, Mitchell, Batra, Zitnick, and Parikh}]{antol2015vqa}
Stanislaw Antol, Aishwarya Agrawal, Jiasen Lu, Margaret Mitchell, Dhruv Batra, C.~Lawrence Zitnick, and Devi Parikh. 2015.
\newblock {VQA}: {V}isual question answering.
\newblock In \emph{The IEEE International Conference on Computer Vision (ICCV)}.

\bibitem[{Cadene et~al.(2019)Cadene, Dancette, Cord, Parikh et~al.}]{cadene2019rubi}
Remi Cadene, Corentin Dancette, Matthieu Cord, Devi Parikh, et~al. 2019.
\newblock Rubi: Reducing unimodal biases for visual question answering.
\newblock In \emph{Advances in Neural Information Processing Systems (NeurIPS)}, pages 839--850.

\bibitem[{Das et~al.(2016)Das, Agrawal, Zitnick, Parikh, and Batra}]{das2016human}
Abhishek Das, Harsh Agrawal, C~Lawrence Zitnick, Devi Parikh, and Dhruv Batra. 2016.
\newblock Human attention in visual question answering: Do humans and deep networks look at the same regions?
\newblock In \emph{Conference on Empirical Methods on Natural Language Processing (EMNLP)}.

\bibitem[{Das et~al.(2017)Das, Agrawal, Zitnick, Parikh, and Batra}]{das2017human}
Abhishek Das, Harsh Agrawal, Larry Zitnick, Devi Parikh, and Dhruv Batra. 2017.
\newblock Human attention in visual question answering: Do humans and deep networks look at the same regions?
\newblock \emph{Computer Vision and Image Understanding (CVIU)}, 163:90--100.

\bibitem[{Gan et~al.(2017)Gan, Li, Li, Sun, and Gong}]{gan2017vqs}
Chuang Gan, Yandong Li, Haoxiang Li, Chen Sun, and Boqing Gong. 2017.
\newblock Vqs: Linking segmentations to questions and answers for supervised attention in vqa and question-focused semantic segmentation.
\newblock In \emph{Proceedings of the IEEE International Conference on Computer Vision}, pages 1811--1820.

\bibitem[{Goyal et~al.(2017)Goyal, Khot, Summers-Stay, Batra, and Parikh}]{goyal2017making}
Yash Goyal, Tejas Khot, Douglas Summers-Stay, Dhruv Batra, and Devi Parikh. 2017.
\newblock Making the {V} in {VQA} matter: Elevating the role of image understanding in visual question answering.
\newblock In \emph{Proceedings of the IEEE Conference on Computer Vision and Pattern Recognition (CVPR)}, volume~1, page~3.

\bibitem[{Grand and Belinkov(2019)}]{grand2019adversarial}
Gabriel Grand and Yonatan Belinkov. 2019.
\newblock Adversarial regularization for visual question answering: Strengths, shortcomings, and side effects.
\newblock In \emph{Proceedings of the Second Workshop on Shortcomings in Vision and Language}, pages 1--13, Minneapolis, Minnesota. Association for Computational Linguistics (ACL).

\bibitem[{Hudson and Manning(2019)}]{Hudson2019GQAAN}
Drew~A Hudson and Christopher~D Manning. 2019.
\newblock {GQA}: A new dataset for real-world visual reasoning and compositional question answering.
\newblock In \emph{Proceedings of the IEEE Conference on Computer Vision and Pattern Recognition (CVPR)}, pages 6700--6709.

\bibitem[{Huk~Park et~al.(2018)Huk~Park, Anne~Hendricks, Akata, Rohrbach, Schiele, Darrell, and Rohrbach}]{huk2018multimodal}
Dong Huk~Park, Lisa Anne~Hendricks, Zeynep Akata, Anna Rohrbach, Bernt Schiele, Trevor Darrell, and Marcus Rohrbach. 2018.
\newblock Multimodal explanations: Justifying decisions and pointing to the evidence.
\newblock In \emph{Proceedings of the IEEE Conference on Computer Vision and Pattern Recognition (CVPR)}, pages 8779--8788.

\bibitem[{Johnson et~al.(2017)Johnson, Hariharan, van~der Maaten, Fei-Fei, Zitnick, and Girshick}]{johnson2017clevr}
Justin Johnson, Bharath Hariharan, Laurens van~der Maaten, Li~Fei-Fei, C~Lawrence Zitnick, and Ross Girshick. 2017.
\newblock Clevr: A diagnostic dataset for compositional language and elementary visual reasoning.
\newblock In \emph{Proceedings of the IEEE Conference on Computer Vision and Pattern Recognition (CVPR)}, pages 1988--1997. IEEE.

\bibitem[{Kafle and Kanan(2017)}]{kafle2017analysis}
Kushal Kafle and Christopher Kanan. 2017.
\newblock An analysis of visual question answering algorithms.
\newblock In \emph{Proceedings of the IEEE International Conference on Computer Vision (ICCV)}, pages 1983--1991. IEEE.

\bibitem[{Kafle et~al.(2018)Kafle, Price, Cohen, and Kanan}]{kafle2018dvqa}
Kushal Kafle, Brian Price, Scott Cohen, and Christopher Kanan. 2018.
\newblock {DVQA}: Understanding data visualizations via question answering.
\newblock In \emph{Proc. IEEE Conference on Computer Vision and Pattern Recognition (CVPR)}, pages 5648--5656.

\bibitem[{Kafle et~al.(2019)Kafle, Shrestha, and Kanan}]{kafle2019challenges}
Kushal Kafle, Robik Shrestha, and Christopher Kanan. 2019.
\newblock Challenges and prospects in vision and language research.
\newblock \emph{Frontiers in Artificial Intelligence}.

\bibitem[{Kafle et~al.(2017)Kafle, Yousefhussien, and Kanan}]{kafle2017data}
Kushal Kafle, Mohammed Yousefhussien, and Christopher Kanan. 2017.
\newblock Data augmentation for visual question answering.
\newblock In \emph{Proceedings of the 10th International Conference on Natural Language Generation (INLG)}, pages 198--202.

\bibitem[{Ramakrishnan et~al.(2018)Ramakrishnan, Agrawal, and Lee}]{ramakrishnan2018overcoming}
Sainandan Ramakrishnan, Aishwarya Agrawal, and Stefan Lee. 2018.
\newblock Overcoming language priors in visual question answering with adversarial regularization.
\newblock In \emph{Advances in Neural Information Processing Systems (NeurIPS)}, pages 1541--1551.

\bibitem[{Ren et~al.(2015)Ren, He, Girshick, and Sun}]{ren2015faster}
Shaoqing Ren, Kaiming He, Ross Girshick, and Jian Sun. 2015.
\newblock Faster {R-CNN}: Towards real-time object detection with region proposal networks.
\newblock In \emph{Advances in Neural Information Processing Systems (NeurIPS)}.

\bibitem[{Selvaraju et~al.(2019)Selvaraju, Lee, Shen, Jin, Ghosh, Heck, Batra, and Parikh}]{selvaraju2019taking}
Ramprasaath~R Selvaraju, Stefan Lee, Yilin Shen, Hongxia Jin, Shalini Ghosh, Larry Heck, Dhruv Batra, and Devi Parikh. 2019.
\newblock Taking a hint: Leveraging explanations to make vision and language models more grounded.
\newblock In \emph{Proceedings of the IEEE International Conference on Computer Vision (ICCV)}, pages 2591--2600.

\bibitem[{Shrestha et~al.(2019)Shrestha, Kafle, and Kanan}]{shrestha2019ramen}
Robik Shrestha, Kushal Kafle, and Christopher Kanan. 2019.
\newblock Answer them all! toward universal visual question answering models.
\newblock In \emph{The IEEE Conference on Computer Vision and Pattern Recognition (CVPR)}.

\bibitem[{Welch(1938)}]{welch1938significance}
Bernard~L Welch. 1938.
\newblock The significance of the difference between two means when the population variances are unequal.
\newblock \emph{Biometrika}, 29(3/4):350--362.

\bibitem[{Wu and Mooney(2019)}]{wu2019self}
Jialin Wu and Raymond Mooney. 2019.
\newblock Self-critical reasoning for robust visual question answering.
\newblock In \emph{Advances in Neural Information Processing Systems (NeurIPS)}, pages 8601--8611.

\end{thebibliography}
